# Multi-view Fuzzy Representation Learning with Rules based Model

Wei Zhang, Zhaohong Deng, *Senior Member, IEEE*, Te Zhang, Kup-Sze Choi, Shitong Wang

*Abstract*—**Unsupervised multi-view representation learning has been extensively studied for mining multi-view data. However, some critical challenges remain. On the one hand, the existing methods cannot explore multi-view data comprehensively since they usually learn a common representation between views, given that multi-view data contains both the common information between views and the specific information within each view. On the other hand, to mine the nonlinear relationship between data, kernel or neural network methods are commonly used for multi-view representation learning. However, these methods are lacking in interpretability. To this end, this paper proposes a new multi-view fuzzy representation learning method based on the interpretable Takagi-Sugeno-Kang (TSK) fuzzy system (MVRL_FS). The method realizes multi-view representation learning from two aspects. First, multi-view data are transformed into a high-dimensional fuzzy feature space, while the common information between views and specific information of each view are explored simultaneously. Second, a new regularization method based on $L_{2,1}$-norm regression is proposed to mine the consistency information between views, while the geometric structure of the data is preserved through the Laplacian graph. Finally, extensive experiments on many benchmark multi-view datasets are conducted to validate the superiority of the proposed method.**

*Index Term*s **– multi-view data, representation learning, fuzzy system, common information, specific information.**

## I. INTRODUCTION

With the advancement of data acquisition technologies, data of different types of features in real-world applications can be readily collected from different sources. For example, different features can be extracted from the audio and image data in videos. Similarly, different features can be extracted from textual data of different languages. These feature types can be considered as descriptions of the subject matter from different views. Compared with single-view data that provides features from one view, multi-view data enable more comprehensive understanding and thus more accurate modeling. Hence, multi-view learning has received great attention in recent years in order to mine multi-view data efficiently. Unsupervised multi-view representation learning is among the hottest research topics in multi-view learning. It concerns the learning of the representation of multi-view data so as to improve the discriminability of the data without using label information and thereby enhancing the robustness of subsequently constructed models [1]. Existing unsupervised

multi-view representation learning algorithms can be classified into the following two categories [2]:

1) *Multi-view alignment based methods*: this category of methods aligns different views to maximize the distributions of the views consistent. The framework of the methods is shown in Fig. 1(a). Some of the representative methods are discussed below. Kan et al. used canonical correlation analysis (CCA) to achieve view alignment by maximizing the correlation between different views [3]. However, this method is only applicable to data with two views and can only explore the linear relationship between data. To address these two problems, Luo et al. proposed a new multi-view representation learning method based on CCA and tensor learning [4]. In addition, a deep CCA method is also proposed in recent years to obtain high-level correlations between data from multiple views [5]. Besides these CCA based methods, methods based on distance and similarity are also proposed. For example, using partial least squares, Li et al. learned the orthogonal projection matrix of two views and minimized the distance of the projected representation [6]. To mine the high-level correlations between the views, Yu et al. proposed a new cross-modal alignment method based on deep neural networks [7]. Zhang et al. learned the new representations of each view by using the similarity of data and aligned them by joint kernel matching [8].

2) *Multi-view fusion based methods*: This category of methods extracts the common information between views and transforms it into a new representation. The framework of the methods is shown in Fig. 1(b). In this category, matrix factorization is a commonly used technique. For example, Liu et al. used non-negative matrix factorization to extract the common representation between views [9]. Considering that the constraints of non-negative matrix factorization are too strict, Luo et al. proposed a new multi-view representation learning method based on semi-negative matrix factorization [10]. Besides, there are also methods using concept factorization [11, 12] and unconstrained matrix factorization [13, 14]. Another commonly used technique is self-representation learning, which assumes that multi-view data share a common self-representation matrix, such as the method proposed by Zhang et al. [15]. To mine the nonlinear relationships within data, Cao et al. used self-representation learning to extract a common representation in the kernel space [16]. Zheng et al. extract a common representation by introducing self-representation learning and degeneration mapping model [17]. In addition, deep neural network based methods have been proposed in

This work was supported in part by the National key R & D plan under Grant (2022YFE0112400), the Chinese Association for Artificial Intelligence (CAAI)-Huawei MindSpore Open Fund under Grant CAAIXSJLJJ-2021-011A, the NSFC under Grant 62176105, the Six Talent Peaks Project in Jiangsu Province under Grant XYDXX-056, the Hong Kong Research Grants Council (PolyU 152006/19E). (Corresponding author: Zhaohong Deng).

W. Zhang, Z. Deng, S. T. Wang are with the School of Artificial Intelligence and Computer Science, Jiangnan University and Jiangsu Key Laboratory of Media Design and Software Technology, Wuxi 214122, China (e-mail: 7201607004@stu.jiangnan.edu.cn; dengzhaohong@jiangnan.edu.cn, wxwangst@aliyun.com).

Te Zhang is with the Lab for Uncertainty in Data and Decision Making (LUCID), School of Computer Science, University of Nottingham, Nottingham, NG81BB, UK (e-mail: te.zhang@nottingham.ac.uk).

K. S. Choi is with The Centre for Smart Health, the Hong Kong Polytechnic University, Hong Kong (e-mail: thomasks.choi@polyu.edu.hk).

recent years. For example, Srivastava et al. learned a representation for each view based on a deep Boltzmann machine and fused them into one representation at the final layer [18]. To remove the irrelevant information between views as much as possible, Wan et al. introduced the information bottleneck theory into deep neural networks and proposed a new deep multi-view representation learning method [19].

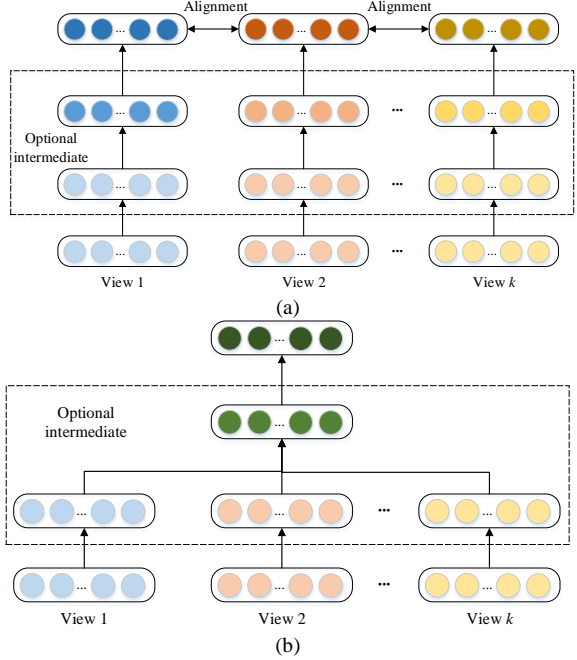

Fig. 1 Two main multi-view representation learning frameworks: (a) the typical framework of multi-views alignment based methods; (b) the typical framework of multi-view fusion based methods.

Although these two categories of methods have achieved great success, they still face critical challenges. First, in multi-view data, there exist not only the common information shared by different views, but also a set of specific information within each view. However, most of the existing methods discussed above only focus on the former and ignore the latter. Second, kernel methods and deep neural networks, that are commonly used to mine the nonlinear relationship between data from multiple views, lack transparency and interpretability. It is also difficult to select a suitable kernel function or network structure for representation learning. Therefore, a more effective and efficient multi-view representation learning method is needed.

Takagi-Sugeno-Kang (TSK) fuzzy system is a data-driven system comprised of IF-THEN fuzzy rules that can create models with strong learning capability as well as good transparency interpretability. It has been successfully applied in various fields [20-22]. In recent years, TSK fuzzy system-based multi-view learning methods have been proposed to achieve better interpretability and deal with uncertainty in multi-view learning [23]. For example, Jiang et al. proposed a multi-view TSK fuzzy system by introducing collaborative learning and maximum margin learning criteria [24]. By learning the hidden view and integrating it with the visible views for modeling, a multi-view TSK fuzzy system leveraging the cooperation between visible and hidden views was proposed [25]. To address the problem in incomplete multi-view classification problem, Zhang et al. integrated common hidden view learning and missing view imputation as one process. Then, they combined common hidden view and imputed multi-view data to construct an incomplete multi-view model [26]. Moreover, a transductive multi-view TSK fuzzy system modeling method was proposed to deal with scenarios where labeled multi-view data were limited [27]. The method learned both the model and the pseudo label simultaneously, and used matrix factorization to further improve the learning abilities of the model. Although these methods have achieved promising performance, they are all proposed for multi-view classification tasks and their performance depends heavily on the learning ability of the specific fuzzy system. There are almost no studies investigating fuzzy system modeling techniques to improve the transparency of multi-view representation learning. In fact, design of TSK fuzzy system based, unsupervised multi-view representation learning methods that can be applied generally to different application scenarios are a challenging task.

To this end, a new multi-view representation learning with TSK fuzzy system (MVRL_FS) is proposed in this paper. First, a multioutput TSK fuzzy system is used as a model for feature extraction. Nonlinear transformation is realized by constructing fuzzy mapping with the antecedent part of the TSK fuzzy system, which does not require the selection of kernel function or deep neural network structure, and can better preserve the information in the original multi-view data [28]. Then, we realize linear dimensionality reduction by learning the consequent parameters of the TSK fuzzy system for each view. In this process, to mine the common information between the views and the specific information within each view simultaneously, we decompose the traditional consequent parameters into the common and the specific parts. In addition, the consistent information mining mechanism, along with $L_{2,1}$-norm regularization regression, is proposed so that the common consequent parameters of different views would tend to be consistent. Finally, a Laplacian graph method and a maximum entropy mechanism are also introduced to preserve the topological structure of multi-view data and to balance the importance of different views, respectively.

The main contributions of this paper are summarized as follows:

1) TSK fuzzy system is the first time introduced into multi-view representation learning, and a new unsupervised multi-view representation learning method is proposed to explore common information between the views and specific information within each view simultaneously.

2) A new multi-view consistent information mining method is proposed with the aid of $L_{2,1}$-norm regularized regression, which greatly improves the discrimination of the learned representation.

3) The effectiveness of the proposed method is validated by comprehensive experimental studies.

The remainder of this paper is organized as follows. The related work is briefly described in Section II. The proposed method is described in detail in Section III. Experimental studies on various datasets are reported in Section IV. Finally, the conclusions and prospects are given in Section V.

Table I Summary of notations frequently used in this paper

| Notation | Description |
|----------|-------------|
| $\mathbf{x}$ | Input feature vector |
| $\mathbf{x}_g$ | Mapped feature vector of input vector $\mathbf{x}$ by the fuzzy rules |
| $\mathbf{x}^v$ | Input feature vector of the $v$th view |
| $\mathbf{X}^v$ | Matrix combining all input vectors of the $v$th view |
| $\mathbf{X}_g^v$ | Matrix combining all the mapped feature vectors of the $v$th view |
| $\mathbf{p}_g$ | Consequent parameters of the single-output TSK fuzzy system |
| $\mathbf{p}_g^v$ | Consequent parameters of the $v$th view single-output TSK fuzzy system |
| $\mathbf{P}^v$ | Consequent parameters of the $v$th view multi-output TSK fuzzy system |
| $\mathbf{P}_c^v$ | Common part consequent parameters of the $v$th view |
| $\mathbf{P}_s^v$ | Specific part consequent parameters of the $v$th view |
| $\mathbf{Z}_c^v$ | Common representation of the $v$th view |
| $\mathbf{Z}_s^v$ | Specific representation of the $v$th view |
| $\mathbf{L}^v$ | Laplacian matrix of the $v$th view |
| $\mathbf{B}$ | Mapping matrix |
| $w^v$ | Weight of the $v$th view |
| $\alpha, \gamma, \beta, \delta$ | regularization parameters |

## II. RELATED WORK

In this section, we first review the existing multi-view representation learning methods that perform nonlinear transformation, and then introduce TKS fuzzy system which is the fundamental of the proposed method. The frequently used notations in this paper are listed in Table I.

### A. Multi-view Representation Learning Methods with Nonlinear Transformation

In recent years, kernel function and deep neural networks have received increasing attention for multi-view learning, based on which a large amount of multi-view representation learning methods have been proposed. Both approaches first perform nonlinear transformation and then project the original multi-view data into the high-dimensional space, followed by linear dimensionality reduction.

For kernel function based multi-view learning, CCA was used by Fukumizu et al. to propose KCCA which exploited the nonlinear relationships between the data of multiple views [29]. Chen et al. proposed a new multi-view representation learning method (JLMVC) [30] to mine the nonlinear relationships between data using a kernel-induced mapping and automatically learn a reasonable weight for each view. Zhang et al. learned the low-rank common representation between views in the kernel space while exploring the effect of different ranks to the model by adopting the weighted Schatten p-paradigms [31]. By using a custom kernel function, De Sa et al. integrated common information between views to obtain a common representation [32]. Similarly, Houthuys et al. used a weighted kernel CCA to learn a common representation between views and obtained the solution using primal-dual optimization [33].

On the other hand, for neural network-based multi-view representation learning method, Andrew et al. proposed the new DeepCCA with deep neural networks to obtain a more extensive association between views [5]. Wang et al. proposed a multi-view representation learning method based on adversarial autoencoder, which aimed at learning a common and compact representation between views [34]. Wang et al.

proposed a multi-view representation learning method based on Deep Neural Networks (DNN) and CCA [35]. Based on autoencoder and low-rank tensor constraint, Zheng et al. extracted a common representation and proposed a multi-view representation learning method [36]. For multi-view graph data, Hassani et al. proposed a multi-view contrastive representation learning method based on Graph Neural Network (GNN) [37]. Similarly, Shao et al proposed a multi-view representation learning for heterogeneous graph data by introducing GNN [38]. Huang et al. proposed a multi-view deep spectral representation learning method by combining deep networks with spectral methods [39].

Although these multi-view representation learning methods with nonlinear transformation have shown promising performance, most of them only focus on the common information between views and ignore the specific information within each view. For this reason, this paper proposes a new multi-view representation learning method that simultaneously mines the common information between views and the specific information within each view. In addition, due to the poor interpretability of kernel function and deep neural network, the transparent and interpretable TSK fuzzy system is adopted as the basis model to develop the corresponding multi-view representation learning method.

### B. Takagi-Sugeno-Kang Fuzzy System

TSK fuzzy system is an intelligent model based on fuzzy set and fuzzy logic [40]. It can construct a model based on expert experience or following a data-driven approach [41]. Compared with traditional machine learning model, TSK fuzzy system has the characteristics of good accuracy and interpretability.

Given a TSK fuzzy system, the $k$th fuzzy rule is often defined as follows:

$$IF\ x_1\ is\ A_1^k \wedge \cdots \wedge\ x_d\ is\ A_d^k$$
$$THEN\ f^k(\mathbf{x}) = p_0^k + p_1^k x_1 + \cdots + p_d^k x_d \quad (1)$$

where $k = 1,2,\ldots,K$, $K$ is the number of rule, $\mathbf{x} = [x_1, x_2, \ldots, x_d] \in R^{1 \times d}$ is the input vector, $d$ is the number of features of $\mathbf{x}$, $f^k(\mathbf{x})$ is the output of the $k$th rule, $A_j^k$ is a fuzzy set associated with the $j$th feature and the $k$th rule, and $\wedge$ is a fuzzy conjunction operator. Unlike the crisp set where the membership is 0 or 1, the membership in the fuzzy set can be any value between 0 and 1. The membership can be calculated using membership functions, which can be defined according to different application scenarios. In the absence of domain knowledge, the Gaussian function is usually used as the membership function, which is defined as follows:

$$\mu_{A_j^k}(x_j) = exp\left(-\left(x_j - e_{k,j}\right)^2 / 2q_{k,j}\right) \quad (2a)$$

where parameters $e_{k,j}$ and $q_{k,j}$ are the center and width of the Gaussian function, respectively. In classical TSK fuzzy system, the two parameters are also called the antecedent parameters. They can be estimated with different strategies, where clustering techniques are typically used, e.g. Fuzzy C-Means Clustering (FCM) [42]. However, since random initialization is involved in FCM, the results may be unstable. To overcome this shortcoming, more stable clustering algorithms such as Var-Part [43] can be adopted.

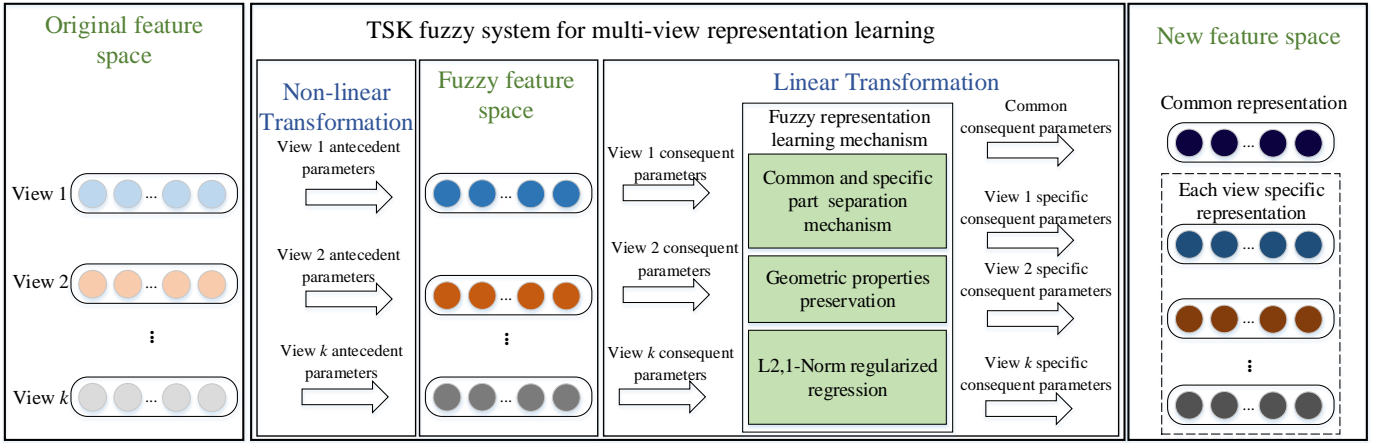

Fig. 2. The framework of the proposed multi-view fuzzy representation learning.

With the known antecedent parameters, the membership value of each feature of the corresponding fuzzy set $A_j^k$ can be calculated by (2a). Using multiplication as the conjunction operator, the firing level of the $k$th rule for each input vector can be calculated using (2b). The normalized form of (2b) is given in (2c). The final output of TSK fuzzy system is then given by the weighted average of $f^k(\mathbf{x})$, as shown in (2d).

$$\mu^k(\mathbf{x}) = \prod_{j=1}^d \mu_{A_j^k}(x_j) \qquad (2b)$$

$$\tilde{\mu}^k(\mathbf{x}) = \frac{\mu^k(\mathbf{x})}{\sum_{i=1}^K \mu^i(\mathbf{x})} \qquad (2c)$$

$$y = \sum_{k=1}^K \tilde{\mu}^k(\mathbf{x}) f^k(\mathbf{x}) \qquad (2d)$$

Once the antecedent parameters are estimated, the outputs of the TSK fuzzy system in (2d) can be expressed in the form of linear regression in a new fuzzy feature space as follows:

$$y = \mathbf{x}_g \mathbf{p}_g \qquad (3a)$$

where $\mathbf{x}_g$ and $\mathbf{p}_g$ are defined as follows:

$$\mathbf{x}_e = [1, \mathbf{x}] \in R^{1 \times (d+1)} \qquad (3b)$$

$$\tilde{\mathbf{x}}^k = \tilde{\mu}^k(\mathbf{x}) \mathbf{x}_e \in R^{1 \times (d+1)} \qquad (3c)$$

$$\mathbf{x}_g = [\tilde{\mathbf{x}}^1, \tilde{\mathbf{x}}^2, \dots, \tilde{\mathbf{x}}^K] \in R^{1 \times K(d+1)} \qquad (3d)$$

$$\mathbf{p}_k = [p_0^k, p_1^k, \dots, p_d^k] \in R^{1 \times (d+1)} \qquad (3e)$$

$$\mathbf{p}_g = [\mathbf{p}_1, \mathbf{p}_2, \dots, \mathbf{p}_K]^T \in R^{K(d+1) \times 1} \qquad (3f)$$

## III. Multi-view Fuzzy Representation Learning

In this section, we propose a new multi-view representation learning method based on fuzzy systems (MVRL_FS) to address the issues mentioned in the introduction. The framework of MVRL_FS is shown in Fig 2. The proposed method can mine the common and specific information of the multi-view data simultaneously, and obtain new representations for both types of information. In addition, MVRL_FS explores the nonlinear relationships between data from multiple views using rules based TSK fuzzy system in order to achieve good performance in both robustness and interpretability.

### A. Multi-view Common and Specific Representation Construction with TSK Fuzzy System

In multi-view representation learning with nonlinear transformation, data is first transformed nonlinearly into a high-dimensional space, which is then linearly reduced to a low-dimensional space. In this paper, the nonlinear transformation is realized by the antecedent parameters of the multi-output TSK fuzzy system, and the linear dimensionality reduction is realized by the consequent parameters of the multi-output TSK fuzzy system.

To construct a TSK fuzzy system for representation learning, the first step is to estimate the antecedent parameters so as to realize the nonlinear transformation from the original space to a high dimensional fuzzy feature space. Different membership functions can be used for the TSK fuzzy system according to specific applications. In this paper, we adopt the Gaussian function as the membership function, and the antecedent parameters are the center and width of the Gaussian function. Specifically, we use the Var-Part clustering algorithm [43] to estimate cluster center $\mathbf{E}^v = [e_{k,j}^v]_{k \times d^v}$ in each view, where $k$ is the number of rules, and $d^v$ is the number of features in the $v$th view. Then the matrix of width parameters in this view, i.e., $\mathbf{Q}^v = [q_{k,j}^v]_{k \times d^v}$ can be estimated as follows:

$$q_{k,j}^v = \sum_{i=1}^N (x_{i,j}^v - e_{k,j}^v)^2 / \sum_{i=1}^K \sum_{i=1}^N (x_{i,j}^v - e_{i,j}^v)^2 \qquad (4)$$

where $j = 1, 2, \dots, d^v$, $x_{i,j}^v$ is the $j$th feature of the $i$th instance in the $v$th view.

When the antecedent parameters of the TSK fuzzy system have been determined, for a given multi-view instance $\{\mathbf{x}^v \in R^{1 \times d^v}, v = 1, 2, \dots, V\}$, it is first transformed into a new fuzzy feature space by (3b) - (3d). The new representation in new the fuzzy feature space can be expressed as follows:

$$\mathbf{x}_g^v = [\tilde{\mathbf{x}}^{1,v}, \tilde{\mathbf{x}}^{2,v}, \dots, \tilde{\mathbf{x}}^{K,v}] \in R^{1 \times K(d^v+1)}, v = 1, 2, \dots, V \quad (5a)$$

Then, we use the consequents of a multi-output TSK fuzzy system as feature transformation $\phi(*)$ to realize linear dimensionality reduction in the constructed fuzzy feature space:

$$\phi(\mathbf{x}^v) = \mathbf{x}_g^v \mathbf{P}^v \qquad (5b)$$

$$\mathbf{P}^v = [\mathbf{p}_g^{1,v}, \mathbf{p}_g^{2,v}, \dots, \mathbf{p}_g^{m,v}] \in R^{K(d^v+1) \times m}, v = 1, 2, \dots, V \quad (5c)$$

The main difference between the single-output TSK fuzzy system in (3a) and the multi-output TSK fuzzy system in (5b)

is that the latter involves multiple group consequent parameters in each fuzzy rule. Therefore, for a given multi-view data, the final representations can be denoted as:

$$\mathbf{Z}^v = \mathbf{X}_g^v \mathbf{P}^v \in R^{N \times m}, v = 1,2,...,V \tag{6a}$$

$$\mathbf{X}_g^v = \left[\mathbf{x}_{g,1}^v; \mathbf{x}_{g,2}^v; ...; \mathbf{x}_{g,N}^v\right] \in R^{N \times K(d^v+1)}, v = 1, 2, ..., V \tag{6b}$$

where $\mathbf{X}_g^v$ is the matrix expression of all instances in $v$th view. $\mathbf{Z}^v$ is the new representation by the transformation using the $m$-dimensional output TSK fuzzy system.

For a multi-view dataset, although the different views describe the same subject, the feature spaces of different views are different. Therefore, there exists not only the common information between views, but also the specific information within each view. However, we can only learn the specific representation of each view through the above method. To address this problem, this paper proposes a new representation learning method based on TSK fuzzy system, which separates the consequent parameter matrix into two parts, i.e., the common part $\mathbf{P}_c^v$ and the specific part $\mathbf{P}_s^v$ as expressed in (7a)-(7c).

$$\mathbf{P}^v = \mathbf{P}_c^v + \mathbf{P}_s^v \tag{7a}$$

$$\mathbf{P}_c^v = \left[\mathbf{p}_{g,c}^{1,v}, \mathbf{p}_{g,c}^{2,v}, ..., \mathbf{p}_{g,c}^{m,v}\right] \in R^{d_g^v \times m} \tag{7b}$$

$$\mathbf{P}_s^v = \left[\mathbf{p}_{g,s}^{1,v}, \mathbf{p}_{g,s}^{2,v}, ..., \mathbf{p}_{g,s}^{m,v}\right] \in R^{d_g^v \times m} \tag{7c}$$

where $d_g^v = K(d^v + 1)$ is the number of features in the fuzzy feature space associated with the $v$th view. $\mathbf{P}_c^v$ and $\mathbf{P}_s^v$ are the consequent parameters of the $m$-dimensional output TSK fuzzy system, which are used to mine the common and specific information, respectively. Then, for a multi-view data, the transformed common and specific representation can be expressed as follows:

$$\mathbf{Z}_c^v = \mathbf{X}_g^v \mathbf{P}_c^v \in R^{N \times m} \tag{8a}$$

$$\mathbf{Z}_s^v = \mathbf{X}_g^v \mathbf{P}_s^v \in R^{N \times m} \tag{8b}$$

### B. Geometric Structure Preservation

How to adapt the learned low-dimensional representation to preserve the geometric structure from the original feature space is a hot research topic in representation learning. The adaptation is an important factor contributing the success of graph-based methods, which allows the low-dimensional representation to contain the geometric structure of the original feature space [44]. Traditional graph-based methods preserve the geometric structure through the following objective function:

$$\min_{\mathbf{Z}} \sum_{i=1}^N \sum_{j=1}^N s_{i,j} \left\| \mathbf{z}_i - \mathbf{z}_j \right\|_2 = tr(\mathbf{Z}^{\mathrm{T}} \mathbf{L} \mathbf{Z}) \tag{9}$$

where $\mathbf{Z} = [\mathbf{z}_1; \mathbf{z}_2; ...; \mathbf{z}_N] \in R^{N \times m}$ is the low-dimensional representation, $\mathbf{S} = \left[s_{i,j}\right]_{N \times N}$ is the similarity matrix which is estimated in the original feature space, $\mathbf{L} = \mathbf{D} - \mathbf{S} \in R^{N \times N}$ is the Laplacian matrix, and $\mathbf{D} \in R^{N \times N}$ is the diagonal matrix with the $i$th diagonal value $d_{i,i} = \sum_{j=1}^N s_{i,j}$.

Since the approach can be only used for single view data, a new collaborative geometric structure preservation method is proposed based on (9) to preserve the geometric structure for the learned common and specific representation simultaneously. The corresponding objective function is defined as follows:

$$\min_{\mathbf{P}_c^v, \mathbf{P}_s^v} \sum_{v=1}^V tr\left( (\mathbf{Z}_c^v + \mathbf{Z}_s^v)^{\mathrm{T}} \mathbf{L}^v \big( (\mathbf{Z}_c^v + \mathbf{Z}_s^v) \big) \right) \tag{10}$$

where $\mathbf{Z}_c^v = \mathbf{X}_g^v \mathbf{P}_c^v$, $\mathbf{Z}_s^v = \mathbf{X}_g^v \mathbf{P}_s^v$, $\mathbf{L}^v = \mathbf{D}^v - \mathbf{S}^v$ is the Laplacian matrix of the $v$th view, $\mathbf{S}^v = \left[s_{i,j}^v\right]_{N \times N}$ is the similarity matrix of $v$th view, and $s_{i,j}^v$ is defined as follows:

$$s_{i,j}^v = \begin{cases} \kappa\left(\mathbf{x}_{g,i}^v, \mathbf{x}_{g,j}^v\right), \text{ if } i\text{-th instance is the } k\text{-nearest} \\ \qquad\qquad \text{neighbor of } j\text{-th instance} \\ \qquad\qquad 0, \text{ otherwise} \end{cases} \tag{11}$$

In (11), $\kappa(*,*)$ is a kernel function and the Gaussian function is used in this paper, and $\mathbf{D}^v$ is the diagonal matrix of $v$th view with $d_{i,i}^v = \sum_{j=1}^N s_{i,j}^v$.

Although the consequent parameters $\mathbf{P}^v$ is divided into the common part $\mathbf{P}_c^v$ and the specific part $\mathbf{P}_s^v$, there is inevitably some redundant information between them since they are learned simultaneously. Moreover, as the discriminative information contained in different views is inconsistent, the discriminability of the learned new representation can be greatly improved if the importance weights of the views can be coordinated. To address these two issues, we introduce the orthogonality constraint and negative Shannon entropy. Accordingly, (10) can be updated as follows:

$$\min_{\mathbf{P}_c^v, \mathbf{P}_s^v, w^v} \sum_{v=1}^V w^v tr\left( (\mathbf{Z}_c^v + \mathbf{Z}_s^v)^{\mathrm{T}} \mathbf{L}^v \big( \mathbf{Z}_c^v + \mathbf{Z}_s^v \big) \right) + \\ \alpha \sum_{v=1}^V \| (\mathbf{Z}_c^v)^{\mathrm{T}}(\mathbf{Z}_s^v) \|_F + \delta \sum_{v=1}^V w^v ln(w^v) \\ s.t. \, w^v \geq 0, \sum_{v=1}^V w^v = 1 \tag{12}$$

where $\|*\|_F$ is the Frobenius norm, $\alpha$ and $\delta$ are regularization parameters, $\sum_{v=1}^V w^v ln^{w^v}$ is the negative Shannon entropy. By introducing Shannon entropy as the regularization term, the weights of different views can be adaptively adjusted [24-26].

### C. Consistency Constraint with $L_{2,1}$ -Norm Regularized Regression

In the above two subsections, we separate the common representation and the specific representation in each view. To further improve the discrimination of the learned common representations, the consistency of information between these common representations needs to be explored. Obviously, if the two original datasets have the same dimension and the distributions are consistent, the new representations of them is also consistent when they are mapped by the same mapping matrix. Therefore, we can make the distributions of common representations $\mathbf{Z}_c^v$ consistent by constructing the following objective function:

$$\min_{\mathbf{P}_c^v, \mathbf{B}} \sum_{v=1}^V \| \mathbf{B} \mathbf{Z}_c^v - \mathbf{H} \|_F + \gamma \sum_{v=1}^V \| \mathbf{B} \|_{2,1} \tag{13}$$

where $\mathbf{B} \in R^{q \times N}$ is the mapping matrix, $L_{2,1}$ regularization is used to ensure the sparsity of $\mathbf{B}$, and $\mathbf{H} \in R^{q \times m}$ is the same low-dimensional representation of all common representations, with $q$ as the number of the features in the low-dimensional space. Compared to finding a suitable value of $q$, we define $\mathbf{H}$ as a $m$-dimensional identity matrix $\mathbf{I}$ and $\mathbf{B}$ as a $m$-dimensional mapping matrix. The effectiveness of this setting is confirmed in our experiments. Thus, (13) can be updated as follows:

$$\min_{\mathbf{P}_c^v, \mathbf{B}} \sum_{v=1}^V \| \mathbf{B} \mathbf{Z}_c^v - \mathbf{I} \|_F + \gamma \| \mathbf{B} \|_{2,1} \tag{14}$$

where $\gamma$ is the regularization parameter. With the help of the regularized regression, (14) is flexible method to ensure the consistency of the common representations of the views.

### D. Overall Objective Function and Optimization

The multi-view representation learning problem can be solved by integrating (12) and (14). Meanwhile, $\sum_{v=1}^{V}\|\mathbf{P}_c^v\|_{2,1}$ and $\sum_{v=1}^{V}\|\mathbf{P}_s^v\|_{2,1}$ are introduced to avoid overfitting. Finally, the overall objective function is defined as follows:

$$\min_{\mathbf{P}_c^v,\mathbf{P}_s^v,\mathbf{B},w^v}\sum_{v=1}^{V}w^v tr\left((\mathbf{Z}_c^v+\mathbf{Z}_s^v)^{\mathrm{T}}\mathbf{L}^v(\mathbf{Z}_c^v+\mathbf{Z}_s^v)\right)+$$
$$\alpha\sum_{v=1}^{V}\|(\mathbf{Z}_c^v)^{\mathrm{T}}(\mathbf{Z}_s^v)\|_F+\beta\sum_{v=1}^{V}\|\mathbf{B}\mathbf{Z}_c^v-\mathbf{I}\|_F+\gamma\|\mathbf{B}\|_{2,1}+$$
$$\gamma\sum_{v=1}^{V}\|\mathbf{P}_c^v\|_{2,1}+\gamma\sum_{v=1}^{V}\|\mathbf{P}_s^v\|_{2,1}+\delta\sum_{v=1}^{V}w^v ln(w^v)$$
$$s.t.\,w^v\geq 0,\sum_{v=1}^{V}w^v=1 \tag{15}$$

where $\mathbf{Z}_c^v=\mathbf{X}_g^v\mathbf{P}_c^v$, $\mathbf{Z}_s^v=\mathbf{X}_g^v\mathbf{P}_s^v$. $\alpha$, $\gamma$, $\beta$, $\delta$ are regularization parameters and they are used to control the impact of the corresponding terms. These hyperparameters can be set manually or determined by cross-validation strategy.

The optimization problem in (15) can be solved iteratively. The optimization procedure is described as follows:

1) **Step 1: Update $\mathbf{P}_c^v$**. Fixing all the variables except $\mathbf{P}_c^v$, we obtain the following minimization problem:

$$\min_{\mathbf{P}_c^v}\sum_{v=1}^{V}w^v tr\left(\left(\mathbf{X}_g^v\mathbf{P}_c^v+\mathbf{X}_g^v\mathbf{P}_s^v\right)^{\mathrm{T}}\mathbf{L}^v\left(\mathbf{X}_g^v\mathbf{P}_c^v+\mathbf{X}_g^v\mathbf{P}_s^v\right)\right)+$$
$$\alpha\sum_{v=1}^{V}\left\|\left(\mathbf{X}_g^v\mathbf{P}_c^v\right)^{\mathrm{T}}\left(\mathbf{X}_g^v\mathbf{P}_s^v\right)\right\|_F+\beta\sum_{v=1}^{V}\|\mathbf{B}\mathbf{X}_g^v\mathbf{P}_c^v-\mathbf{I}\|_F+$$
$$\gamma\sum_{v=1}^{V}\|\mathbf{P}_c^v\|_{2,1} \tag{16}$$

Taking the derivative of (16) with respect to $\mathbf{P}_c^v$ and setting it to zero, the update rule for $\mathbf{P}_c^v$ is obtained as:

$$\mathbf{P}_c^v=\left(w^v\mathbf{X}_g^{v,\mathrm{T}}\mathbf{L}^v\mathbf{X}_g^v+\alpha\mathbf{X}_g^{v,\mathrm{T}}\mathbf{X}_g^v\mathbf{P}_s^v\mathbf{P}_s^{v,\mathrm{T}}\mathbf{X}_g^{v,\mathrm{T}}\mathbf{X}_g^v+\right.$$
$$\left.\beta\mathbf{X}_g^{v,\mathrm{T}}\mathbf{B}^{\mathrm{T}}\mathbf{B}\mathbf{X}_g^v+\gamma\mathbf{F}_c\right)^{-1}\left(\beta\mathbf{X}_g^{v,\mathrm{T}}\mathbf{B}^{\mathrm{T}}-w^v\mathbf{X}_g^{v,\mathrm{T}}\mathbf{L}^v\mathbf{X}_g^v\mathbf{P}_s^v\right) \tag{17}$$

where $\mathbf{F}_c$ is a diagonal matrix, and the $i$th diagonal value $\mathbf{F}_{c,i,i}=1/\|\mathbf{P}_{c,i,:}^v\|_2$.

2) **Step 2: Update $\mathbf{P}_s^v$**. Fixing all the variables except $\mathbf{P}_s^v$, we obtain the following minimization problem:

$$\min_{\mathbf{P}_s^v}\sum_{v=1}^{V}w^v tr\left(\left(\mathbf{X}_g^v\mathbf{P}_c^v+\mathbf{X}_g^v\mathbf{P}_s^v\right)^{\mathrm{T}}\mathbf{L}^v\left(\mathbf{X}_g^v\mathbf{P}_c^v+\mathbf{X}_g^v\mathbf{P}_s^v\right)\right)+$$
$$\alpha\sum_{v=1}^{V}\left\|\left(\mathbf{X}_g^v\mathbf{P}_c^v\right)^{\mathrm{T}}\left(\mathbf{X}_g^v\mathbf{P}_s^v\right)\right\|_F+\gamma\sum_{v=1}^{V}\|\mathbf{P}_s^v\|_{2,1} \tag{18}$$

Taking the derivative of (18) with respect to $\mathbf{P}_s^v$ and setting it to zero, the update rule for $\mathbf{P}_s^v$ is obtained as:

$$\mathbf{P}_s^v=\left(w^v\mathbf{X}_g^{v,\mathrm{T}}\mathbf{L}^v\mathbf{X}_g^v+\alpha\mathbf{X}_g^{v,\mathrm{T}}\mathbf{X}_g^v\mathbf{P}_c^v\mathbf{P}_c^{v,\mathrm{T}}\mathbf{X}_g^{v,\mathrm{T}}\mathbf{X}_g^v+\right.$$
$$\left.\gamma\mathbf{F}_s\right)^{-1}\left(-w^v\mathbf{X}_g^{v,\mathrm{T}}\mathbf{L}^v\mathbf{X}_g^v\mathbf{P}_c^v\right) \tag{19}$$

where $\mathbf{F}_s$ is a diagonal matrix, and the $i$th diagonal value $\mathbf{F}_{s,i,i}=1/\|\mathbf{P}_{s,i,:}^v\|_2$.

3) **Step 3: Update $\mathbf{B}$**. Fixing all the variables except $\mathbf{B}$, we obtain the following minimization problem:

$$\min_{\mathbf{B}}\sum_{v=1}^{V}\|\mathbf{B}\mathbf{X}_g^v\mathbf{P}_c^v-\mathbf{I}\|_F+\gamma\|\mathbf{B}\|_{2,1} \tag{20}$$

Taking the derivative of (20) with respect to $\mathbf{B}$ and setting it to zero, the update rule for $\mathbf{B}$ is obtained as:

$$\mathbf{B}=(\mathbf{I}+\gamma\mathbf{F}_b)^{-1}\left(\sum_{v=1}^{V}\mathbf{P}_c^{v,\mathrm{T}}\mathbf{X}_g^{v,\mathrm{T}}\right) \tag{21}$$

where $\mathbf{F}_b$ is also a diagonal matrix, and the $i$th diagonal value $\mathbf{F}_{b,i,i}=1/\|\mathbf{B}_{i,:}\|_2$.

4) **Step 4: Update $w^v$**. Fixing all the variables except $w^v$, we obtain the following minimization problem:

$$\min_{w^v}\sum_{v=1}^{V}w^v tr\left((\mathbf{Z}_c^v+\mathbf{Z}_s^v)^{\mathrm{T}}\mathbf{L}^v(\mathbf{Z}_c^v+\mathbf{Z}_s^v)\right)+$$
$$\delta\sum_{v=1}^{V}w^v ln(w^v)$$
$$s.t.\,w^v\geq 0,\sum_{v=1}^{V}w^v=1 \tag{22}$$

Using Lagrangian optimization method, the update rule for $w^v$ can be obtained as:

$$w^v=\frac{exp\left(-tr\left((\mathbf{Z}_c^v+\mathbf{Z}_s^v)^{\mathrm{T}}\mathbf{L}^v(\mathbf{Z}_c^v+\mathbf{Z}_s^v)\right)/\delta\right)}{\sum_l^V exp\left(-tr\left((\mathbf{Z}_c^l+\mathbf{Z}_s^l)^{\mathrm{T}}\mathbf{L}^l(\mathbf{Z}_c^l+\mathbf{Z}_s^l)\right)/\delta\right)} \tag{23}$$

By solving the optimization problem iteratively using the update rules (17), (19), (21) and (23), the local optimal solution of (15) can be obtained.

Based on the above analyses, the algorithm of the proposed MVRL_FS is given in Algorithm 1. There are three parameters ($\mathbf{P}_c^v$, $\mathbf{P}_s^v$, $w^v$) needed be initialized in Algorithm 1. We randomly initialize $\mathbf{P}_s^v$, $\mathbf{P}_c^v$ and let each view weight $w^v=1/V$.

---

**Algorithm 1 MVRL_FS**

---

**Input:** multi-view data $\{\mathbf{X}^v\}$, $v=1, 2, \ldots, V$, the number of fuzzy rules $K$; number of maximum iterations $T$, the regularization parameters $\alpha$, $\gamma$, $\beta$, $\delta$.
**Output:** $\mathbf{P}_c^v,\mathbf{P}_s^v,\mathbf{B},w^v$.

1: Use the Var-Part clustering algorithm to estimate the antecedent parameters of the TSK fuzzy systems based on data $\mathbf{X}^v$ for different views.
2: Use (3.b) - (3.d) to construct a new dataset $\mathrm{D}_l^v=\{\mathbf{X}_g^v\}$ in the fuzzy feature space generated by fuzzy rules for each view.
3: Initialize $\mathbf{P}_s^v$, $\mathbf{P}_c^v$, $w^v$.
4: for $t=1, 2, \ldots, T$ do
5: Update $\mathbf{B}$ based on (21).
6: for $v=1, 2, \ldots, V$ do
7: Update $\mathbf{P}_c^v$ based on (17).
8: Update $\mathbf{P}_s^v$ based on (19).
9: Update $w^v$ based on (23).
10: end
11: end

---

### E. Complexity Analysis

In MVRL_FS, the feature dimension $m$ of the common and specific representation is usually much less than the number of instances $N$, the feature dimension of the data in the original space $d^v$ and fuzzy feature space $d_g^v$. The time complexities of steps 1 and 2 in Algorithm 1 are $O(2Nd^vK)$ and $O((1+d^v)NK)$ respectively, where $K$ is the number of rules. The time complexity of step 5 is $O\left((d_g^v+m)mNT\right)$. The time complexities of step 7 and step 8 are $O\left((N^2+Nd_g^v+d_g^vm+Nm)d_g^vVT\right)$, where $V$ is the number of views and $T$ is the maximum number of iterations. Finally, the time complexity of step 9 is $O\left((m+N+d_g^v)mNVT\right)$. Therefore, the computational complexity of the overall algorithm is $O(N^2d_g^{v,2}VT)$.

## F. Convergence Analysis

In this subsection, the convergence of Algorithm 1 is shown by proving the theorem below.

*Theorem 1*: The objective function in (15) is bounded. The value of the objective function at each iteration decreases monotonically with Algorithm 1.

*Proof*: we define $J(\mathbf{P}_c^v, \mathbf{P}_s^v, \mathbf{B}, w^v)$ as the objection function in (15). It is obvious that $J(\mathbf{P}_c^v, \mathbf{P}_s^v, \mathbf{B}, w^v) > 0$. Therefore, the objective function in (15) is bounded. Then, in Algorithm 1, the objective function $J(\mathbf{P}_c^v, \mathbf{P}_s^v, \mathbf{B}, w^v)$ has four variables and can be divided correspondingly into four subproblems, each being a convex problem with respect to a variable. Therefore, the optimal solution of the subproblems can be obtained by updating (17), (19), (21) and (23). As a result, the value of $J(\mathbf{P}_c^v, \mathbf{P}_s^v, \mathbf{B}, w^v)$ is decreased at each iteration of Algorithm 1.

## G. New Feature Presentation for Subsequent Modeling Tasks

By using the common and specific representations learned with Algorithm 1, we can construct the new representation as follows:

$$\mathbf{Z} = \left[ \sum_{v=1}^{V} w^v \mathbf{X}_g^v \mathbf{P}_c^v , w^1 \mathbf{X}_g^1 \mathbf{P}_s^1 , w^2 \mathbf{X}_g^2 \mathbf{P}_s^2 , ... , w^V \mathbf{X}_g^V \mathbf{P}_s^V \right] \quad (24)$$

In (24), the first term is the integrated common representation of all views, and the other terms are the weighted specific representations of different views. A small weight means that the effect of the corresponding view is weak. Then, the new representation of the original multi-view dataset is fed into the classification or clustering model, such as SVM or K-means, to perform the subsequent modeling tasks.

## IV. EXPERIMENTAL STUDIES

In this section, the effectiveness of the proposed algorithm is verified. Section IV-A gives the details of the experimental setting, which includes the datasets, the comparison methods, the parameters setting and the evaluation indices. Experimental results and analysis are given in Section IV-B. The ablation studies and statistical analysis are presented in Sections IV-C and IV-D, respectively. The running time and convergence analysis of the proposed algorithm are given in Section IV-E. Finally, Section IV-F presents the interpretability analysis of the model. The code of the proposed MVRL_FS is available at https://github.com/BBKing49/MVRL_FS.

### A. Experimental Settings

#### 1) Datasets

Seven real-world multi-view datasets were adopted as the baselines in our experiments, which are briefly described below. Table II gives the statistics of datasets.

1) Corel [27]: an image dataset containing 1000 instances and 10 classes. SIFT and LBP features were extracted as two views of the experiments.

2) NUS-WIDE [45]: an image dataset containing 1000 instances and 5 classes. CORR and bag of visual words features are extracted as two views for experiments.

3) MF [27]: a handwriting dataset containing 2000 images with 10 classes. Fourier coefficients and Zernike moments features were extracted as two views for experiments.

Table II Statistics of multi-view datasets

| Dataset | Size | Number of Views (Dimensions) | Number of Classes |
|---|---|---|---|
| Corel | 1000 | 2 (256-300) | 10 |
| NUS-WIDE | 1000 | 2 (144-500) | 5 |
| MF | 2000 | 2 (76-47) | 10 |
| MSRCv1 | 240 | 3 (24-256-254) | 7 |
| ORL | 400 | 4 (512-59-864-256) | 40 |
| Leavers | 1600 | 3 (64-64-64) | 100 |
| CCA | 6773 | 3 (20-20-20) | 20 |

Table III Parameters setting of the algorithms

| Algorithms | Parameters and grid search range |
|---|---|
| MSC_IAS | Regularization parameter $\lambda_1$, $\lambda_2$: {1,2, 3, ..., 30}. The dimension of common representation: {100, 200, ..., 500}. |
| LMSC | Regularization parameters $\lambda$: {$10^{-3}$, $10^{-2}$, ..., $10^2$, $10^3$}. The dimension of common representation: {10, 20, ..., 90, 100}. |
| CSMSC | Regularization parameters $\lambda_1$, $\lambda_2$: {$10^{-3}$, $10^{-2}$, ..., $10^2$, $10^3$}. |
| NMFCC | Regularization parameters $\alpha$, $\beta$, $\gamma$, $\mu$: {$10^{-3}$, $10^{-2}$, ..., $10^2$, $10^3$}. The dimension of common representation is set as the number of clusters. |
| DiMSC | Regularization parameters $\lambda_1$, $\lambda_2$: {$10^{-3}$, $10^{-2}$, ..., $10^2$, $10^3$}. |
| TCCA | Regularization parameter $\lambda$: {$10^{-3}$, $10^{-2}$, ..., $10^2$, $10^3$}, The dimension of common representation: {10, 20, ..., 90, 100}. |
| MDcR | Regularization parameters $\lambda$: {$10^{-3}$, $10^{-2}$, ..., $10^2$, $10^3$}. The dimension of common representation: {10, 20, ..., 90, 100}. |
| MVRL_FS | Regularization parameters $\alpha$, $\gamma$, $\beta$, $\delta$: {$2^{-5}$, $2^{-4}$, ..., $2^4$, $2^5$}. The dimension of common and specific representation is set as the number of classes. The number of rules is 3. |

4) MSRCv1 [46]: a dataset containing 210 images and 7 classes, i.e., cows, trees, buildings, airplanes, faces, and cars. CMT, LBP, and GENT features were extracted as three views for experiments.

5) ORL [47]: a dataset containing 400 facial images of different people. CENTRIST, LBP, GIST, and HOG features were extracted as four views for experiments.

6) Leavers [48]: a plant image dataset containing 100 categories of plants. Shape, Texture and Margin features were extracted as three views for experiments.

7) CCA [49]: a YouTube video dataset containing 20 semantic classes. SIFT, STIP, and MFCC features were extracted as three views for experiments.

#### 2) Methods for Comparison

Seven state-of-the-art multi-view representation learning methods were compared with the proposed method. Four of them (MSC_IAS, LMSC, CSMSC and NMFCC) are based on linear transformation, and the remaining three (DiMSC, TCCA and MDcR) are based on nonlinear transformation. The detailed descriptions of these methods are given below.

1) MSC_IAS [50]: this method extracts a common representation between views based on matrix factorization with similarity constraints.

2) LMSC [15]: this method extracts a common representation between views based on self-representation learning.

3) CSMSC [51]: this method extracts a common representation and a set specific representations based on self-representation learning for multi-view data, which are combined to form a new representation.

Table IV. Performance multi-view representation learning methods on seven datasets

| Datasets | Algorithms | NMI | ACC | Purity |
|---|---|---|---|---|
| Corel | MSC_IAS | 0.2577±0.0081 | 0.3297±0.0180 | 0.3763±0.0171 |
| | LMSC | 0.2892±0.0021 | 0.3870±0.0155 | 0.4253±0.0110 |
| | CSMSC | 0.1990±0.0030 | 0.3077±0.0095 | 0.3363±0.0087 |
| | NMFCC | 0.2706±0.0016 | 0.3460±0.0079 | 0.3940±0.0020 |
| | DiMSC | 0.2449±0.0063 | 0.3422±0.0073 | 0.3690±0.0089 |
| | TCCA | 0.3129±0.0109 | 0.3887±0.0302 | 0.4413±0.0289 |
| | MCDR | 0.2793±0.0117 | 0.3543±0.0093 | 0.3997±0.0104 |
| | MVRL_FS (ours) | **0.3196±0.0034** | **0.3907±0.0080** | **0.4493±0.0075** |
| NUS-WIDE | MSC_IAS | 0.0901±0.0039 | 0.3403±0.0064 | 0.3697±0.0060 |
| | LMSC | 0.1088±0.0016 | 0.3893±0.0029 | 0.3957±0.0023 |
| | CSMSC | 0.0387±0.0068 | 0.2847±0.0081 | 0.2990±0.0052 |
| | NMFCC | 0.1020±0.0070 | 0.3746±0.0081 | 0.3857±0.0110 |
| | DiMSC | 0.1277±0.0018 | 0.4350±0.0035 | 0.4350±0.0035 |
| | TCCA | 0.1286±0.0083 | 0.3920±0.0312 | 0.4080±0.0178 |
| | MCDR | 0.1123±0.0024 | 0.3870±0.0026 | 0.3970±0.0010 |
| | MVRL_FS (ours) | **0.1604±0.0020** | **0.4470±0.0030** | **0.4577±0.0015** |
| MF | MSC_IAS | 0.7053±0.0341 | 0.7310±0.0618 | 0.7423±0.0424 |
| | LMSC | 0.4782±0.0116 | 0.5362±0.0447 | 0.5767±0.0087 |
| | CSMSC | 0.6682±0.0059 | 0.7505±0.0053 | 0.7505±0.0053 |
| | NMFCC | 0.6099±0.0193 | 0.6642±0.0645 | 0.6688±0.0569 |
| | DiMSC | 0.6063±0.0050 | 0.6857±0.0060 | 0.6860±0.0056 |
| | TCCA | 0.7144±0.0000 | 0.7780±0.0000 | 0.7780±0.0000 |
| | MCDR | 0.7087±0.0159 | 0.7486±0.0045 | 0.7582±0.0206 |
| | MVRL_FS (ours) | **0.7571±0.0128** | **0.7738±0.0166** | **0.8012±0.0071** |
| MSRCv1 | MSC_IAS | 0.3263±0.0865 | 0.3381±0.0497 | 0.3714±0.0701 |
| | LMSC | 0.5278±0.0615 | 0.6333±0.0874 | 0.6651±0.0607 |
| | CSMSC | 0.3922±0.0132 | 0.5841±0.0192 | 0.5921±0.0180 |
| | NMFCC | 0.5274±0.0666 | 0.6302±0.0738 | 0.6619±0.0667 |
| | DiMSC | 0.5908±0.0113 | **0.6952±0.0058** | 0.7190±0.0058 |
| | TCCA | 0.5727±0.0666 | 0.6603±0.0674 | 0.6730±0.0667 |
| | MCDR | 0.4479±0.0325 | 0.4603±0.0592 | 0.5238±0.0412 |
| | MVRL_FS (ours) | **0.5911±0.0212** | 0.6794±0.0275 | **0.7222±0.0275** |
| ORL | MSC_IAS | 0.9101±0.0066 | 0.8275±0.0152 | 0.8508±0.0142 |
| | LMSC | 0.9307±0.0116 | 0.8550±0.0241 | 0.8833±0.0184 |
| | CSMSC | 0.8326±0.0176 | 0.7783±0.0230 | **0.8933±0.0030** |
| | NMFCC | 0.8644±0.0022 | 0.7333±0.0101 | 0.7650±0.0050 |
| | DiMSC | 0.9178±0.0051 | 0.8140±0.0145 | 0.8465±0.0170 |
| | TCCA | 0.8208±0.0148 | 0.6000±0.0115 | 0.6583±0.0218 |
| | MCDR | 0.8077±0.0079 | 0.6408±0.0052 | 0.6783±0.0063 |
| | MVRL_FS (ours) | **0.9369±0.0129** | **0.8642±0.0210** | 0.8825±0.0238 |
| Leavers | MSC_IAS | 0.8975±0.0027 | 0.8092±0.0088 | 0.8231±0.0098 |
| | LMSC | 0.7458±0.0066 | 0.5231±0.0220 | 0.5502±0.0138 |
| | CSMSC | 0.8728±0.0114 | 0.7452±0.0183 | 0.7742±0.0149 |
| | NMFCC | 0.9270±0.0095 | 0.7846±0.0229 | 0.8196±0.0208 |
| | DiMSC | 0.8748±0.0076 | 0.7340±0.0230 | 0.7620±0.0224 |
| | TCCA | 0.9305±0.0013 | **0.8213±0.0111** | 0.8399±0.0073 |
| | MCDR | 0.9349±0.0059 | 0.7990±0.0254 | 0.8392±0.0213 |
| | MVRL_FS (ours) | **0.9355±0.0033** | 0.7822±0.0111 | **0.8433±0.0118** |
| CCA | MSC_IAS | 0.1512±0.0026 | 0.1769±0.0028 | 0.2271±0.0041 |
| | LMSC | 0.1104±0.0000 | 0.1537±0.0000 | 0.1049±0.0000 |
| | CSMSC | 0.1653±0.0010 | 0.2138±0.0000 | 0.2438±0.0000 |
| | NMFCC | 0.1428±0.0109 | 0.1807±0.0121 | 0.2145±0.0078 |
| | DiMSC | 0.1208±0.0022 | 0.1508±0.0026 | 0.2007±0.0013 |
| | TCCA | 0.1895±0.0026 | 0.2076±0.0068 | **0.2499±0.0041** |
| | MCDR | 0.1544±0.0085 | 0.1824±0.0067 | 0.2254±0.0084 |
| | MVRL_FS (ours) | **0.2004±0.0021** | **0.2116±0.0054** | 0.2492±0.0031 |

4)NMFCC [52]: this method extracts a common representation between views based on non-negative matrix factorization with orthogonal constraints.

5)DiMSC [16]: this method extracts a common representation between views based on self-representation learning in the kernel space.

6)TCCA [4]: this method aligns multiple views based on CCA incorporating tensor techniques.

7) MDcR [8]: this method uses the projection matrix to align multiple views in the kernel space.

*3) Parameters setting*

The hyperparameters of the algorithms adopted and the corresponding grid search ranges are shown in Table III.

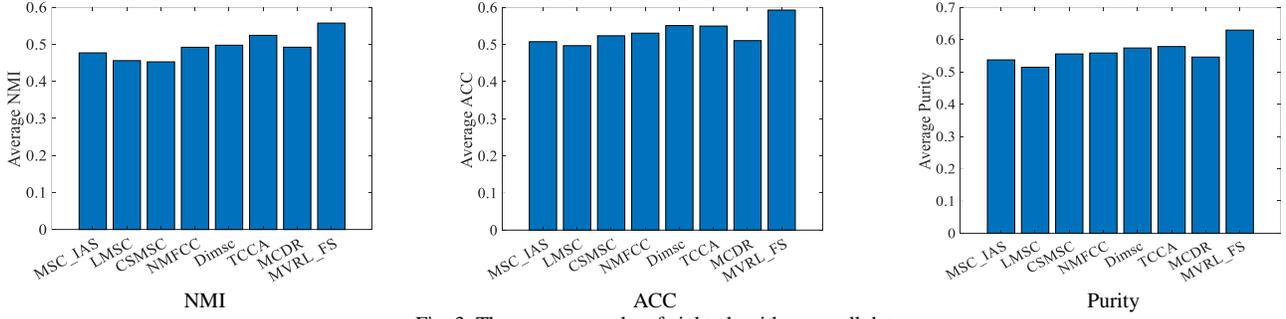

Fig. 3. The average results of eight algorithms on all datasets.

### 4) Performance Evaluation Scheme and Indices

To evaluate the representation learning methods, we follow the approach in [13, 18] and consider clustering as the modeling task. When new representations are obtained by different representation learning methods, clustering is implemented by K-means. Then the clustering results are used to evaluate the performance of the methods. The clustering performance was evaluated using the three commonly used indices NMI, ACC, and Purity [53, 54]. Moreover, each algorithm was executed 20 times with different parameters, and the best results in terms of the mean and the standard deviation of the metrics were recorded for comparison.

### B. Clustering Performance

The performance of the eight methods on the seven datasets is presented in terms of the three evaluation indices in Table IV. The average performance is also show in Fig. 3. From the results, the following observations can be obtained.

1) The proposed MVRL_FS has shown the best performance in most datasets. This indicates the effectiveness of TSK fuzzy system based nonlinear transformation and the mechanism of exploring common and specific representation simultaneously.

2) In most cases, the performance of the linear transformation based multi-view representation methods, i.e., MSC_IAS, LMSC, CSMSC and NMFCC, is much inferior to that of the nonlinear transformation based counterparts, i.e., DiMSC, TCCA, MDcR and MVRL_FS. For example, for the NUS-WIDE dataset, the average NMI values of the nonlinear transformation based methods are 2%-3% higher than that of the linear transformation based methods. For the MF dataset, the average NMI values of nonlinear transformation based methods are 5%-10% higher than that of the linear transformation based methods.

3) As a linear transformation based multi-view representation method, CSMSC has an advantage over other linear methods on most datasets, which indicates that mining the common and specific information between views can effectively improve representation learning ability. For example, for the CCA dataset, the average NMI values of CSMSC are 1%-4% higher than other linear transformation based methods. However, we can see that the performance of CSMSC is still inferior to the proposed MVRL_FS for all datasets, which indicates that only mining the linear common and specific information is insufficient, and nonlinear learning are therefore required.

4) Compared with other multi-view representation learning methods based on nonlinear transformation, the proposed MVRL_FS is highly competitive, especially for the Corel and MF datasets, where MVRL_FS has a 5% higher NMI than other nonlinear transformation based methods. This indicates that TSK fuzzy system is very effective in mining the nonlinear relationships between data. Besides, it also indicates that mining and combining the common and specific information can greatly enhance the discriminability of the learned representation.

### C. Ablation Studies

Ablation studies were conducted to further evaluate the effectiveness of the mechanism of learning both the common and specific representations simultaneously, as well as the ability of the proposed consistency information mining mechanism. We define the method that only learns the common representation as MVRL_FS1 and that ignores the consistency information as MVRL_FS2. The corresponding objective functions are defined respectively in (25) and (26) as follows:

$$\min_{\mathbf{P}_c^v, \mathbf{B}, w^v} \sum_{v=1}^{V} w^v tr\big((\mathbf{Z}_c^v)^\mathrm{T} \mathbf{L}^v (\mathbf{Z}_c^v)\big) + \beta \sum_{v=1}^{V} \|\mathbf{B}\mathbf{Z}_c^v - \mathbf{I}\|_F +$$
$$\gamma \|\mathbf{B}\|_{2,1} + \gamma \sum_{v=1}^{V} \|\mathbf{P}_c^v\|_{2,1} + \delta \sum_{v=1}^{V} w^v ln(w^v)$$
$$\text{s. t.}\, w^v \geq 0, \sum_{v=1}^{V} w^v = 1 \tag{25}$$

$$\min_{\mathbf{P}_c^v, \mathbf{P}_s^v, \mathbf{B}, w^v} \sum_{v=1}^{V} w^v tr\big((\mathbf{Z}_c^v + \mathbf{Z}_s^v)^\mathrm{T} \mathbf{L}^v (\mathbf{Z}_c^v + \mathbf{Z}_s^v)\big) +$$
$$\alpha \sum_{v=1}^{V} \|(\mathbf{Z}_c^v)^\mathrm{T}(\mathbf{Z}_s^v)\|_F + \gamma \|\mathbf{B}\|_{2,1} + \gamma \sum_{v=1}^{V} \|\mathbf{P}_c^v\|_{2,1} +$$
$$\gamma \sum_{v=1}^{V} \|\mathbf{P}_s^v\|_{2,1} + \delta \sum_{v=1}^{V} w^v ln(w^v)$$
$$\text{s. t.}\, w^v \geq 0, \sum_{v=1}^{V} w^v = 1 \tag{26}$$

An alternating optimization scheme is used to solve the objective functions. Table V gives the experimental results of the three versions of the algorithm on all the datasets. It is found that mining both the common and specific information between views is advantageous over mining the common information only in most cases, especially on the MSRCv1 dataset. The ACC and Purity values of MVRL_FS1 are nearly 10% lower than that of MVRL_FS. In addition, it can be seen that the performance of MVRL_FS2 is significantly worse than MVRL_FS and MVRL_FS1 on all the datasets, which indicates the effectiveness of the proposed consistency information mining mechanism.

Table V. The results of MVRL_FS, MVRL_FS1 and MVRL_FS2 on all datasets

| Datasets | MVRL_FS | | | MVRL_FS1 | | | MVRL_FS2 | | |
|---|---|---|---|---|---|---|---|---|---|
| | NMI | ACC | Purity | NMI | ACC | Purity | NMI | ACC | Purity |
| Corel | **0.3196** ±0.0034 | **0.3907** ±0.0080 | **0.4493** ±0.0075 | 0.2918 ±0.0042 | 0.3657 ±0.0042 | 0.4177 ±0.0032 | 0.2562 ±0.0037 | 0.3467 ±0.0072 | 0.3827 ±0.0055 |
| NUS-WIDE | **0.1604** ±0.0020 | 0.4470 ±0.0030 | **0.4577** ±0.0015 | 0.1598 ±0.0033 | **0.4503** ±0.0050 | 0.4503 ±0.0050 | 0.1212 ±0.0005 | 0.3923 ±0.0005 | 0.4053 ±0.0005 |
| MF | **0.7571** ±0.0128 | **0.7738** ±0.0166 | **0.8012** ±0.0071 | 0.7176 ±0.0007 | 0.7480 ±0.0007 | 0.7738 ±0.0007 | 0.5781 ±0.0170 | 0.5788 ±0.0521 | 0.6200 ±0.0485 |
| MSRCv1 | **0.5911** ±0.0212 | **0.6794** ±0.0275 | **0.7222** ±0.0275 | 0.5837 ±0.0001 | 0.5714 ±0.0000 | 0.6517 ±0.0000 | 0.4562 ±0.0281 | 0.4984 ±0.0153 | 0.5746 ±0.0198 |
| ORL | **0.9369** ±0.0129 | **0.8642** ±0.0210 | **0.8825** ±0.0238 | 0.9313 ±0.0191 | 0.8454 ±0.0521 | 0.8742 ±0.0355 | 0.8414 ±0.0112 | 0.6950 ±0.0319 | 0.7308 ±0.0213 |
| Leavers | **0.9355** ±0.0033 | 0.7822 ±0.0111 | **0.8433** ±0.0118 | 0.9205 ±0.0012 | **0.7873** ±0.0013 | 0.8242 ±0.0042 | 0.8326 ±0.0053 | 0.6610 ±0.0314 | 0.6992 ±0.0248 |
| CCA | **0.2004** ±0.0021 | 0.2116 ±0.0054 | **0.2492** ±0.0031 | 0.1988 ±0.0020 | **0.2233** ±0.0067 | 0.2436 ±0.0019 | 0.1691 ±0.0026 | 0.1958 ±0.0019 | 0.2252 ±0.0036 |

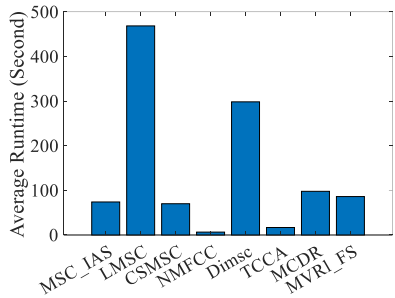

Fig. 4. The average runtime of the algorithms on the eight datasets.

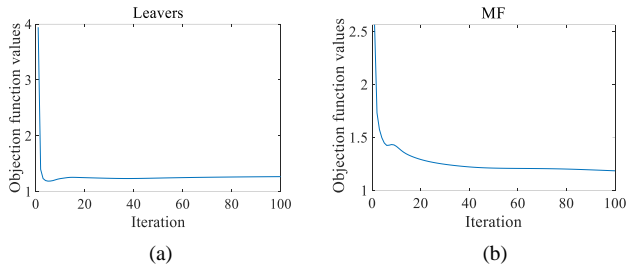

Fig. 5: Convergence curves of MVRL_FS on Leavers and MF datasets.

Table VI. Friedman Test based on NMI

| Algorithm | Ranking | p-value | Null Hypothesis |
|---|---|---|---|
| MVRL_FS | 1 | | |
| TCCA | 3 | | |
| MCDR | 4.4286 | | |
| Dimsc | 5 | 0.001139 | Rejected |
| NMFCC | 5.2857 | | |
| LMSC | 5.4286 | | |
| MSC_IAS | 5.5714 | | |
| CSMSC | 6.2857 | | |

## D. Runtime and Convergence Analyses

We compare the average runtime of the eight algorithms on all datasets. The results are shown in Fig. 4. It can be seen that the running time of MVRL_FS is average, better than three while having no advantage over the other four – NMFCC, TCCA, MSC_IAS, and CSMSC. Nevertheless, MVRL_FS outperforms the other algorithms in terms of the three main evaluation indices as shown in Table IV and Fig. 3.

Furthermore, we conduct convergence analysis based on the Leavers and MF datasets. Fig. 5 plots the values of the objective function over time (iteration steps). We find that the value decreases rapidly and convergence can be reached within 60 steps of iteration.

## E. Statistical Analysis

Friedman test [55] and the post-hoc Holm test [56] were conducted to evaluate the statistical significance of the performance advantage of the proposed MVRL_FS in clustering over the seven algorithms under comparison.

First, we conducted the Friedman test according to [55]. The null hypothesis was that there was no difference in performance between the eight algorithms. When the $p$-value of the test was less than 0.05, the null hypothesis was rejected. Table VI gives the results of the Friedman test based on NMI. Tables S1-S2 in the *Supplementary Materials* section give the results based on Purity and ACC. As shown in Tables VI and S1-S2, all the $p$-values are less than 0.05. Therefore, the difference in performance between the algorithms is significant. Among them, MVRL_FS ranks first, suggesting that it is the best algorithms.

Next, the post-hoc Holm test [56] was conducted to verify whether MVRL_FS performs significantly better than the other seven methods. The results based on NMI is given in Table VII, and those based on Purity and ACC can be found in Tables S3-S4 in the *Supplementary Materials* section. All the three tables show that the performance difference between MVRL_FS and six methods is significant, except for TCCA. However, the results in Table IV and Fig. 3 indicate that MVRL_FS still outperforms TCCA to a certain extent.

## F. Interpretability Analysis

The interpretability of fuzzy systems is mainly attributed to the rule-based knowledge expression and the fuzzy inference mechanism. TSK fuzzy system is commonly used in regression or classification tasks. In this paper, it is considered as a multi-view representation learning model. Compared with kernel based and deep network based methods, TSK fuzzy system based method makes the process of representation learning more interpretable, and the representation learning process can be interpreted using a set of fuzzy rules.

Table VII. Post-hoc Test based on NMI (reject hypothesis if $p$-value $<0.05$)

| $i$ | Algorithms | $z = (R_0 - R_i)/SE$ | $p$-value | Holm$=\alpha/i$ | Null Hypothesis |
|---|---|---|---|---|---|
| 7 | CSMSC/MVRL_FS | 4.037031 | 0.000054 | 0.007143 | Rejected |
| 6 | MSC_IAS/MVRL_FS | 3.491486 | 0.00048 | 0.008333 | Rejected |
| 5 | LMSC/MVRL_FS | 3.382377 | 0.000719 | 0.01 | Rejected |
| 4 | NMFCC/MVRL_FS | 3.273268 | 0.001063 | 0.0125 | Rejected |
| 3 | Dimsc/MVRL_FS | 3.05505 | 0.00225 | 0.016667 | Rejected |
| 2 | MCDR/MVRL_FS | 2.618615 | 0.008829 | 0.025 | Rejected |
| 1 | TCCA/MVRL_FS | 1.527525 | 0.12663 | 0.05 | Not Rejected |

Table VIII. Rule base generated for common representation on Corel dataset in the SIFT view

**The rule base of the TSK fuzzy system for the common representation**

**Rule 1**:
**IF**: the 1th feature is High and
the 2th feature is High and
…, and
the 300th feature is High.
**Then**: the 1th output is -0.0038-0.0011$x_1$+0.0072$x_2$+…+0.0298$x_{300}$ and
the 2th output is 0.0074+0.0034$x_1$+0.0048$x_2$+…+0.0151$x_{300}$ and
…, and
the 10th output is -0.0045-0.0065$x_1$-0.0069$x_2$+…+0.0085$x_{300}$
**Rule 2**:
**IF**: the 1th feature is Low and
the 2th feature is Middle and
…, and
the 300th feature is Low.
**Then**: the 1th output is -0.0038-0.0017$x_1$+0.0088$x_2$+…+0.0385$x_{300}$ and
the 2th output is 0.0053+0.0017$x_1$+0.0067$x_2$+…+0.0175$x_{300}$ and
…, and
the 10th output is -0.0037-0.0049$x_1$-0.0076$x_2$+…+0.0158$x_{300}$
**Rule 3**:
**IF**: the 1th feature is Middle and
the 2th feature is Low and
…, and
the 300th feature is Middle.
**Then**: the 1th output is -0.0040-0.0019$x_1$+0.0085$x_2$+…+0.038$x_{300}$ and
the 2th output is 0.0175+0.0065$x_1$+0.0026$x_2$+…+0.0172$x_{300}$ and
…, and
the 10th output is 0.0158-0.0045$x_1$-0.0056$x_2$+…+0.0040$x_{300}$

Table IX. Rule base generated for specific representation on Corel dataset in the SIFT view

**The rule base of the TSK fuzzy system for the specific representation**

**Rule 1**:
**IF**: the 1th feature is High and
the 2th feature is High and
…, and
the 300th feature is High.
**Then**: the 1th output is 0.0035+0.0008$x_1$-0.0072$x_2$+…-0.0062$x_{300}$ and
the 2th output is -0.0071-0.0029$x_1$-0.005$x_2$+…-0.0039$x_{300}$ and
…, and
the 10th output is 0.0044+0.0062$x_1$+0.0066$x_2$+…+0.0047$x_{300}$
**Rule 2**:
**IF**: the 1th feature is Low and
the 2th feature is Middle and
…, and
the 300th feature is Low.
**Then**: the 1th output is 0.0036+0.0015$x_1$-0.0083$x_2$+…+0.0064$x_{300}$ and
the 2th output is -0.0054-0.0059$x_1$-0.0058$x_2$+…-0.0055$x_{300}$ and
…, and
the 10th output is 0.0036+0.0048$x_1$+0.0071$x_2$+…-0.0015$x_{300}$
**Rule 3**:
**IF**: the 1th feature is Middle and
the 2th feature is Low and
…, and
the 300th feature is Middle.
**Then**: the 1th output is 0.0036+0.0011x1-0.0081x2+…-0.0129$x_{300}$ and
the 2th output is -0.0055-0.0059x1-0.0020x2+…-0.0057$x_{300}$ and
…, and
the 10th output is 0.0043+0.0053x1+0.0073x2+…+0.0107$x_{300}$

In the proposed method, the multiple outputs of the TSK fuzzy system represent the new features. The rules for feature transformation can be formulated as follows:

$$IF \ x_1 \ is \ A_1^k(e_1^k, q_1^k) \ \wedge \cdots \wedge \ x_d \ is \ A_d^k(e_d^k, q_d^k)$$
$$THEN \ f^k(\mathbf{x}) = [p_0^{k,1} + p_1^{k,1}x_1 + \cdots + p_d^{k,1}x_d, p_0^{k,2} + p_1^{k,2}x_1 + \cdots + p_d^{k,2}x_d, \ldots, p_0^{k,m} + p_1^{k,m}x_1 + \cdots + p_d^{k,m}x_d] \ k = 1, \ldots, K$$
(27)

Each fuzzy set $A_j^k(e_j^k, q_j^k)$ in the antecedent, associated with the $j$th dimension in the $k$th rule, can be interpreted with a linguistic description. In our experiment, the number of fuzzy rules is set as three, i.e., there are three clustering centers for each dimension. According to the order of the clustering center value, the linguistic terms of the corresponding fuzzy sets are *Low*, *Medium* and *High*, respectively. Note that the linguistic descriptions given here are only a possible way to explain the IF-part of the fuzzy rule. Other descriptions can be used depending on the application scenario.

As an example, we use the SIFT view in the Corel dataset to illustrate the interpretability of the proposed MVRL_FS, where Gaussian membership function is adopted. Since the dataset of this view has 300 dimensions, we only use three of them in the illustration, the first, second and last (i.e., the 300th). The membership functions and the possible linguistic explanation of each fuzzy set are shown in Fig. S1 in the *Supplementary Materials* section. For the first dimension (first row in Fig. S1), the center and variance of the first fuzzy set are 0.0858 and 4.1820, where the center value 0.0858 ranks first among the three centers (i.e., 0.0858, 0.0458, and 0.0760). Therefore, this fuzzy set is expressed as *High*. The other dimensions can be analyzed in the same way. Once all fuzzy sets are described with linguistic terms, the fuzzy systems can be explained by fuzzy rules. The fuzzy rule base is given in Tables VIII and IX, where the common and specific representation learning processes are described as a set of rules, respectively.

## V. CONCLUSION

In this paper, we propose the new multi-view representation learning method based on fuzzy system MVRL_FS. It realizes nonlinear transformation by constructing fuzzy mapping with the antecedent part of the TSK fuzzy system, and then explores the common and specific information in the fuzzy feature space by learning the consequence part of the TSK fuzzy system. In addition, $L_{2,1}$-norm regularization regression is further proposed to enhance the consistency of the common representation of each view. At the same time, a Laplacian graph method and a maximum entropy mechanism are introduced to preserve the topological geometric structure and to balance the importance of different views, respectively. The experimental results show that MVRL_FS has a better performance than many existing algorithms.

Although MVRL_FS has achieved promising performance, there is still some room for improvement. First, the similarity matrix in the geometric structure preservation needs to be learned in advance and cannot be integrated into the process of representation learning. Second, MVRL_FS is a two-step method, i.e., MVRL_FS cannot integrate the representation learning with the subsequent task, such as classification or clustering. When the subsequent task is fixed, one-step method can be investigated. Third, MVRL_FS considers the common and specific information as a linear relationship, but in some complex scenarios, it may be difficult to separate common and specific information simply by using linear method. The above issues will be addressed in depth in our future work.

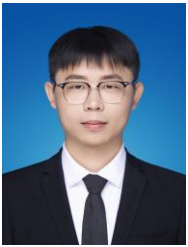

**Wei Zhang** is currently pursuing the Ph.D. degree in the School of Artificial Intelligence and Computer Science, Jiangnan University, Wuxi, China.

His research interests include computational intelligence, machine learning, interpretable artificial intelligence, fuzzy modeling and their applications.

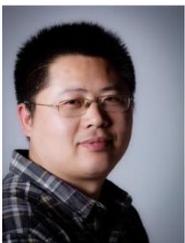

**Zhaohong Deng** (M'12-SM'14) received the B.S. degree in physics from Fuyang Normal College, Fuyang, China, in 2002, and the Ph.D. degree in information technology and engineering from Jiangnan University, Wuxi, China, in 2008.

He is currently a Professor with the School of Artificial Intelligence and Computer Science, Jiangnan University. He has visited the University of California-Davis and the Hong Kong Polytechnic University for more than two years. His current research interests include interpretable intelligence, uncertainty in artificial intelligence and their applications. He has authored or coauthored more than 100 research papers in international/ national journals.

Dr. Deng has served as an Associate Editor or Guest Editor of several international Journals, such as IEEE Trans. Emerging Topics in Computational Intelligence, *Neurocomputing*, and so on.

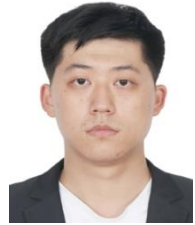

**Te Zhang** received the M.S. degree in Software Engineering from the School of Digital Media, Jiangnan University, Wuxi, China, in 2018. He is currently pursuing the Ph.D. degree in the Lab for Uncertainty in Data and Decision Making (LUCID), School of Computer Science, University of Nottingham, Nottingham, UK. He has been a Research Assistant in the Computer Science and Engineering Department, Southern University of Science and Technology, Shenzhen, China.

His current research interests include fuzzy systems, explainable artificial intelligence and causal discovery.

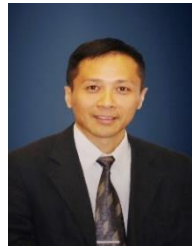

**Kup-Sze Choi** (M'97) received the Ph.D. degree in computer science and engineering from the Chinese University of Hong Kong, Hong Kong in 2004.

He is currently an Professor at the School of Nursing, Hong Kong Polytechnic University, Hong Kong, and the Director of the Centre for Smart Health. His research interests include virtual reality, artificial intelligence, and their applications in medicine and healthcare.

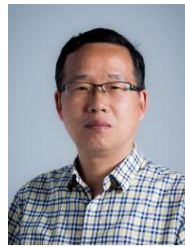

**Shitong Wang** received the M.S. degree in computer science from Nanjing University of Aeronautics and Astronautics, Nanjing, China, in 1987.

His research interests include artificial intelligence, neuro-fuzzy systems, pattern recognition, and image processing. He has published more than 100 papers in international/ national journals and has authored 7 books.